\documentclass[letterpaper, 10pt]{ieeeconf}
\IEEEoverridecommandlockouts

\usepackage{amsmath} 
\usepackage{graphicx}
\usepackage{siunitx}
\usepackage{nicefrac} 
\usepackage{cite}
\usepackage{subfig}
\usepackage[colorlinks = true, urlcolor = blue,citecolor = black, linkcolor = black]{hyperref}
\usepackage{adjustbox}
\usepackage{balance}
\usepackage[font=footnotesize]{caption}
\usepackage{threeparttable}
\usepackage{soul}
\usepackage{verbatim}
\usepackage{url}
\usepackage{amssymb}
\usepackage{mathtools}
\usepackage{xparse}
\usepackage[percent]{overpic}

\usepackage{geometry}
\makeatletter
\newcommand{\raisemath}[1]{\mathpalette{\raiseMath{#1}}}
\newcommand{\raiseMath}[3]{\raisebox{#1}[0pt][0pt]{$#2#3$}}
\makeatother

\NewDocumentCommand{\qbar}{O{0.5pt} O{-6.7pt}}{
	\ensuremath{\mathrlap{\raisemath{#2}{\hspace*{#1}{\mathchar'26\mkern-9mu}}} q}%
}

\NewDocumentCommand{\pbar}{O{-1.5pt} O{-6.7pt}}{
	\ensuremath{\mathrlap{\raisemath{#2}{\hspace*{#1}{\mathchar'26\mkern-9mu}}} p}%
}
\newcommand{\bs}[1]{\boldsymbol{#1}}  
\newcommand{\ts}[1]{\text{#1}} 

\renewcommand{\baselinestretch}{0.8598}  

\title{\LARGE \bf
Control of Flying Robotic Insects: A Perspective and Unifying Approach}

\author{A.~A.~Calder\'on, Y.~Chen, X.~Yang, L.~Chang, X.-T.~Nguyen, E.~K.~Singer, and N.~O.~P\'erez-Arancibia\\
\vspace{-4ex}
\thanks{This work was partially supported by the \textit{National Science Foundation} (NSF) through NRI Award\,1528110, the Chilean National Office of Scientific and Technological Research (CONICYT) through a graduate fellowship to A. A. Calder\'on, and the USC Viterbi School of Engineering through a start-up fund to N. O. P\'erez-Arancibia.}
\thanks{The authors are with the Department of Aerospace and Mechanical Engineering, University of Southern California (USC), Los Angeles, CA 90089-1453, USA (e-mail: {\tt aacalder@usc.edu; chen061@usc.edu; xiufeng@usc.edu; longlonc@usc.edu; xuantrun@usc.edu;\,eksinger@usc.edu;\,perezara@usc.edu}).}%
}

\begin{document}

\maketitle
\thispagestyle{empty}
\pagestyle{empty}

\begin{abstract}
We discuss the problem of designing and implementing controllers for insect-scale \textit{flapping-wing micro air vehicles} (FWMAVs), from a unifying perspective and employing two different experimental platforms; namely, a Harvard RoboBee-like two-winged robot and the four-winged USC Bee\textsuperscript{+}. Through experiments, we demonstrate that a method that employs quaternion coordinates for attitude control, developed to control quadrotors, can be applied to drive both robotic insects considered in this work. The proposed notion that a generic strategy can be used to control several types of artificial insects with some common characteristics was preliminarily tested and validated using a set of experiments, which include position- and attitude-controlled flights. We believe that the presented results are interesting and valuable from both the research and educational perspectives.
\end{abstract}

\vspace{-1ex}
\section{Introduction}
\label{SEC01}
\vspace{-0.5ex}
Insect-sized \textit{flapping-wing micro air vehicles} (FWMAVs) have the potential to become useful tools in search and rescue, exploration in hazardous environments, assisted agriculture and reconnaissance. In the past few years, new mm-to-cm robotic designs inspired by animals with the ability to fly and hover with high maneuverability and efficiency, such as bees, flies and hummingbirds, have been developed. Research on FWMAVs is motivated not only by immediate applications but also reflects the desire to study the not-yet-fully-understood aerodynamic mechanisms employed by flapping-wing insects to fly. In addition, there still are numerous research challenges to be overcome in terms of design, fabrication and materials as the structural and functional components (transmissions, actuators, rotational springs, et cetera) of vehicles of this scale can not be developed or studied using the paradigms applicable to human-scale robotic systems. For example, friction plays a much more significant role at the mm-scale than at the m-scale and, therefore, most classical mechanisms are ineffective in the development of microrobots. In this case, to deal with issues of this type, we adopted the \textit{smart composite microstructures} (SCM) approach~\cite{wood2003microrobotics}, which uses lightweight carbon-fiber composites with high tensile strength for structural purposes, electroactive materials such as piezoelectric ceramics that enable high-precision and high-frequency actuation, and flexible joints and transmission mechanisms made from polyimide film (Kapton).

Overall, a combination of factors make the control of microrobots in general, and FWMAVs in particular, very challenging. Here, we discuss two control cases from a unifying perspective; as platforms, we employ a 75-mg two-winged FWMAV (similar to the Harvard RoboBee~\cite{ma2012design,ma2013controlled}) and a 95-mg four-winged FWMAV (a USC Bee\textsuperscript{+} prototype~\cite{yang2019bee}). These controllable designs were developed upon numerous ideas, results, mechanisms and prototypes produced by several groups in the past twenty years, including the butterfly-like ornithopter in~\cite{tanaka2005flight}, the transmission in~\cite{fearing2000wing,yan2002wing}, the passive wing-pitching mechanism in~\cite{wood2007design} and the two-winged Harvard Fly in~\cite{wood2008microrobot}. The two wings of the prototype in~\cite{wood2008microrobot} are driven by a single central bimorph piezoelectric actuator; the configuration employed to achieve both the first liftoff~\cite{wood2008microrobot} and the first controlled vertical flight~\cite{perez2011first} at the insect-scale. However, the utilization of a single actuator prevented these robots from becoming fully autonomous as severe underactuation makes the system uncontrollable. Further improvements in robotic design and microfabrication enabled researchers to create the split-actuator mechanism~\cite{ma2012design}, composed by two bimorph piezoelectric actuators, that drives the controllable robot in~\cite{ma2013controlled} (RoboBee). Theoretically, this design is fully controllable; in practice, it is able to reliably generate controllable pitch and roll torques but not consistent yaw torques due to an insufficient actuation speed, which becomes an issue in the presence of fabrication errors. To address the problem of insufficient yaw control authority, in~\cite{yang2019bee} we introduced a new four-winged design that is driven by two pairs of twinned actuators. We show that despite their intrinsic differences, robots driven by two and four wings can be controlled employing the same high-level approach.
\begin{figure*}[t!]
\vspace{2ex}
\begin{center}
\includegraphics{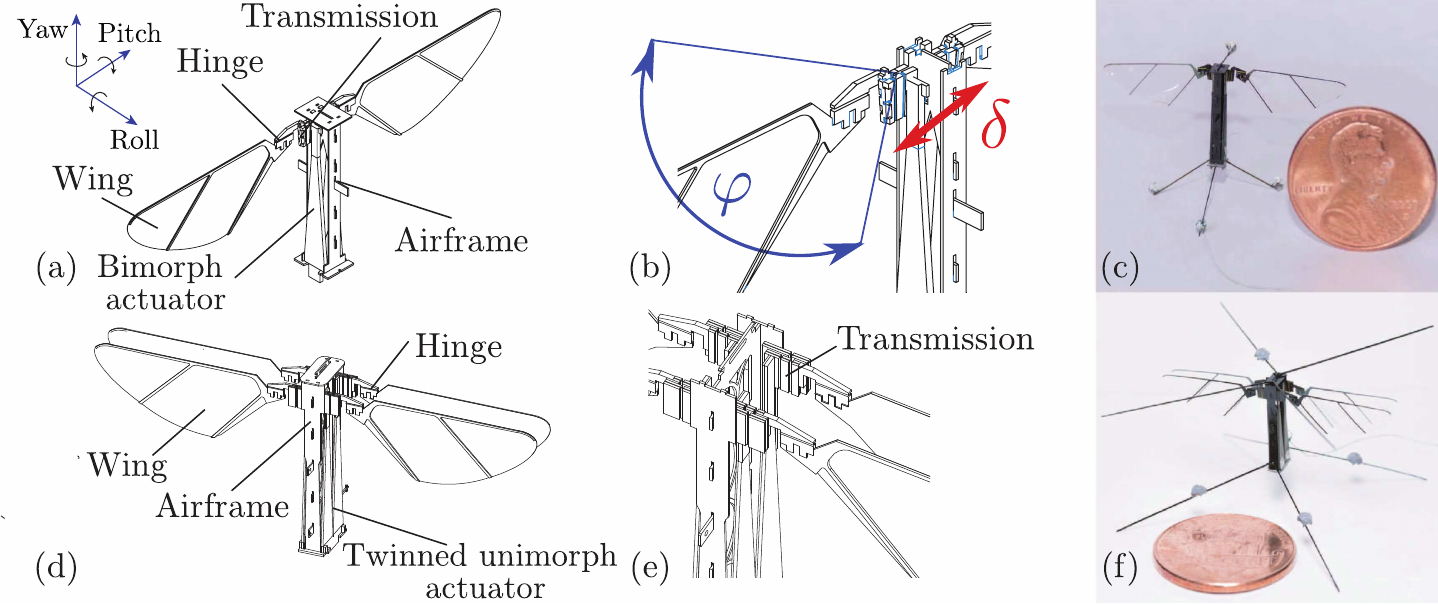}
\end{center}
\vspace{-1ex}
\caption{Schematic diagrams and photographs of the two robotic designs used in this work. (a)~This figure illustrates the mechanical design of the RoboBee. (b)~A flexure transmission converts the approximately linear motion generated by a bimorph piezoelectric actuator, $\delta$, into a flapping angle, $\varphi$. (c)~Photograph of a RoboBee prototype. Four retroreflective 5-mg markers for motion tracking are attached. A U.S. penny indicates the scale. (d)~This figure illustrates the mechanical design of Bee\textsuperscript{+}. (e)~A zoomed-in view of the flapping mechanisms. Two pairs of unimorph actuators drive four wings to flap independently through four transmissions. (f)~A photograph of a Bee\textsuperscript{+} prototype. This robot has a mass of \SI{95}{\milli\gram} and a wingspan of \SI{33}{\milli\meter}. Protective spars and legs are installed on the top and bottom of the robot for crash-protection and landing during flight experiments. Four 5-mg retroreflective markers for motion tracking are attached as well. A U.S. penny indicates the scale. \label{FIG01}}
\vspace{-2ex}
\end{figure*}

In recent years, significant progress on the control of FWMAVs has been reported. In \cite{perez2015model}, a model-free heuristic method is introduced to control a RoboBee prototype, which achieved straight vertical flight and hovering. However, due to its simple structure and the direct use of the local attitude coordinate, this method only achieves local stability with a relatively small region of attraction. In~\cite{ma2012design}, a RoboBee prototype is demonstrated to perform controlled flights and hover at desired positions, employing a simple \textit{linear time-invariant} (LTI) controller that is analyzed using Lyapunov methods. In~\cite{chirarattananon2014adaptive}, a sliding-mode-based adaptive controller is introduced to compensate for uncertainties in the value of the moment of inertia and constant torque disturbances affecting the RoboBee during flight. In~\cite{fuller2019four}, an attitude control approach for a new cross-shaped four-winged flying robot was successfully proposed and validated. The attitude control methods in \cite{ma2012design,chirarattananon2014adaptive,fuller2019four} are based on the idea of aligning the direction of the thrust force with a desired reference without considering the regulation of the yaw angle. 
  
In~\cite{yang2019bee}, the problem of controlling a Bee\textsuperscript{+} prototype is addressed with a position-and-attitude strategy inspired by methods developed to control quadrotors~\cite{chen2017lyapunov}. Namely, using quaternion analysis, an attitude control principle is derived from Euler's rotation theorem by aligning two attitudes in $SO(3)$ instead of two vectors in $\mathbb{R}^3$~\cite{murray2017mathematical}. In this work, we show that both the RoboBee and Bee\textsuperscript{+} can be modeled as single rigid bodies, and that essentially the same method can be employed to control quadrotors, two-winged FWMAVs and four-winged FWMAVs according to a unifying approach. The rest of the paper is organized as follows. Section\,\ref{SEC02} discusses the design of the microrobots; Section\,\ref{SEC03} describes the microfabrication processes; Section\,\ref{SEC04} presents a comparison from the aerodynamics perspective of the RoboBee and Bee\textsuperscript{+}; Section\,\ref{SEC05} explains the control algorithms used to perform flight experiments; and experimental results are presented and discussed in Section\,\ref{SEC06}. Lastly, Section\,\ref{SEC07} draws conclusions and states directions for future research.

\vspace{-1ex}
\section{Design}
\label{SEC02}
\vspace{-0.5ex}
The design of the Bee\textsuperscript{+} is based on that of the RoboBee, with the major innovation being the use of four wings instead of two, each with its own corresponding actuator and transmission mechanism. The actuators of Bee\textsuperscript{+} are designed, fabricated and assembled in twinned pairs.

\vspace{-1ex}
\subsection{RoboBee Design}
\label{SEC02A}
\vspace{-0.5ex}
The RoboBee-like robot used in this research weighs $\sim$75\,mg, uses two wings for lift generation, and has a total wingspan of 35\,mm (see Figs.\,\ref{FIG01}(a)--\ref{FIG01}(c)). In addition to the wings, as shown in Fig.\,\ref{FIG01}(a), the robot is composed of a central airframe; two bimorph piezoelectric actuators; two transmissions that translate the approximately linear motions generated by the actuators into rotational motions, as shown in Fig.\,\ref{FIG01}(b); and two flexure hinges which enable the wings to passively rotate. The resulting robot's configuration resembles those of two-winged insects such as flies, and four-winged insects with coupled forewings and hindwings such as bees. 

In the cases considered in this paper, the wing motion can be decomposed into three modes; namely, flapping, pitching and stroke-plane deviation. For a robot resting on a flat horizontal plane as shown in Fig.\,\ref{FIG01}(c), flapping is defined as the rotation of a wing about the vertical axis that intersects the wing's root (angle $\varphi$ in Fig.\,\ref{FIG01}(b)); pitching is the rotation of the wing about its leading edge; and the stroke-plane deviation is the rotation of the wing about the axis that intersects the wing's root and is parallel to the robot's roll axis as defined in Fig.\,\ref{FIG01}(a). To generate the actuation required to regulate all three rotational motions independently, at least three actuators per wing are necessary, which is a significant challenge with current design and fabrication technologies. The RoboBee design solved this issue by using a single bimorph actuator, per wing, to generate the flapping motion; while, an elastic hinge mechanism allows the wing to pitch passively as a result of its interaction with the surrounding fluid. As part of the fabrication process, the total instantaneous vertical force generated by the prototype in Fig.\,\ref{FIG01}(c) was measured using the micro-force sensor presented in~\cite{singer2018clip}. According to the collected data, this robot produces an average vertical force of 137\,mg (1343\,$\mu$N), including aerodynamic lift and inertial components. This value is similar to the 139\,mg reported in~\cite{ma2012design} for the Harvard prototype.

\vspace{-1ex}
\subsection{Bee\textsuperscript{+} Design}
\label{SEC02B}
\vspace{-0.5ex}
Compared to the two-winged RoboBee, the additional pair of wings of Bee\textsuperscript{+} brings three main advantages: (1)~the control authority is increased, which is advantageous for controlling flight maneuvers; (2)~simple aerodynamic analyses indicate that the four-winged configuration and flapping mode of Bee\textsuperscript{+} damp the rotational disturbances that typically affect the yaw motion of RoboBee prototypes; and (3)~the life-expectancy of Bee\textsuperscript{+}s is expected to be longer than those of RoboBees as the mechanical loading on each wing is lower. The prototype in Figs.\,\ref{FIG01}(d)--\ref{FIG01}(f) weighs 95\,mg and has a wingspan of 33\,mm. 

The key innovation that made the development of Bee\textsuperscript{+} possible was the invention of a new actuation mechanism composed of two pairs of twinned unimorph actuators. This configuration allowed us to overcome the main challenges associated with the design and fabrication of four-winged insect-sized flying robots. From the design perspective, weight reduction is a crucial element to increase the lift-to-weight ratio of the robot. As at this scale the main weight contributors are the actuators, the design of Bee\textsuperscript{+} is not functionally feasible if piezoelectric bimorphs are used. To see this point, note that one pair of twinned unimorph actuators (28\,mg), such as those in the prototype of Fig.\,\ref{FIG01}(f), weighs only 3\,mg more than a single bimorph (25\,mg), used in the prototype in Fig.\,\ref{FIG01}(c). As a result, Bee\textsuperscript{+} (95\,mg) is not significantly heavier than the RoboBee (75\,mg) and much lighter than the four-winged robot (143\,mg) driven by four bimorph actuators that is reported in\,\cite{fuller2019four}. 

From the fabrication perspective, integrating multiple actuators into a functional insect-sized system is challenging because the performance and functionality of microrobots greatly depend on the uniformity of fabrication and precision of assembly. Regarding this aspect, the twinned actuator design enabled us to devise a highly precise fabrication procedure in which the number of misalignments and asymmetries introduced during assembly is drastically reduced because each twinned actuator pair is fabricated monolithically from a single piece of composite material. Furthermore, Bee\textsuperscript{+} requires only five wires to power its two pairs of twinned unimorph actuators; in contrast, powering four bimorph actuators requires a minimum of six wires~\cite{fuller2019four}.

\vspace{-1ex}
\section{Fabrication}
\label{SEC03}
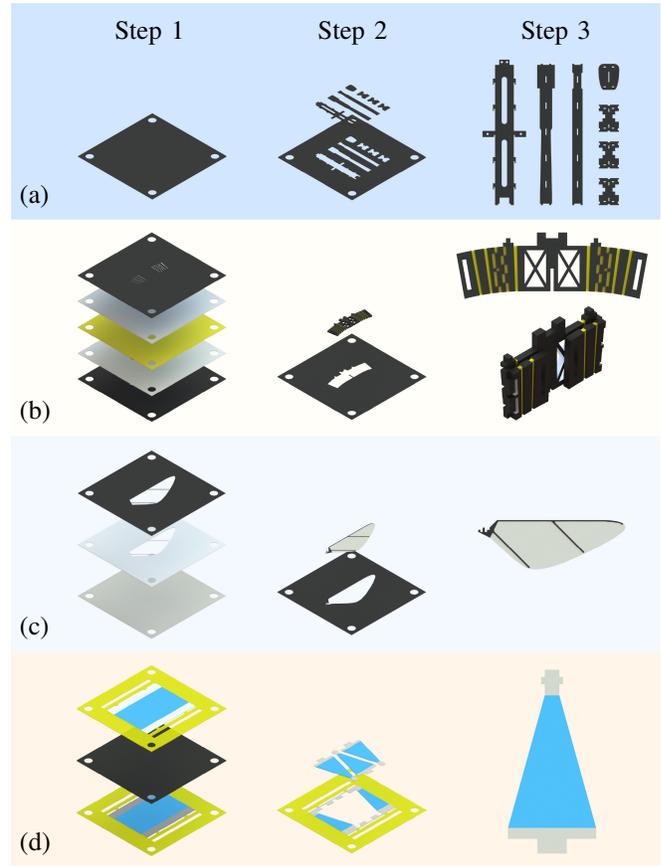
\begin{figure}[t!]
\vspace{2ex}
\begin{overpic}[width=\linewidth,tics=10]{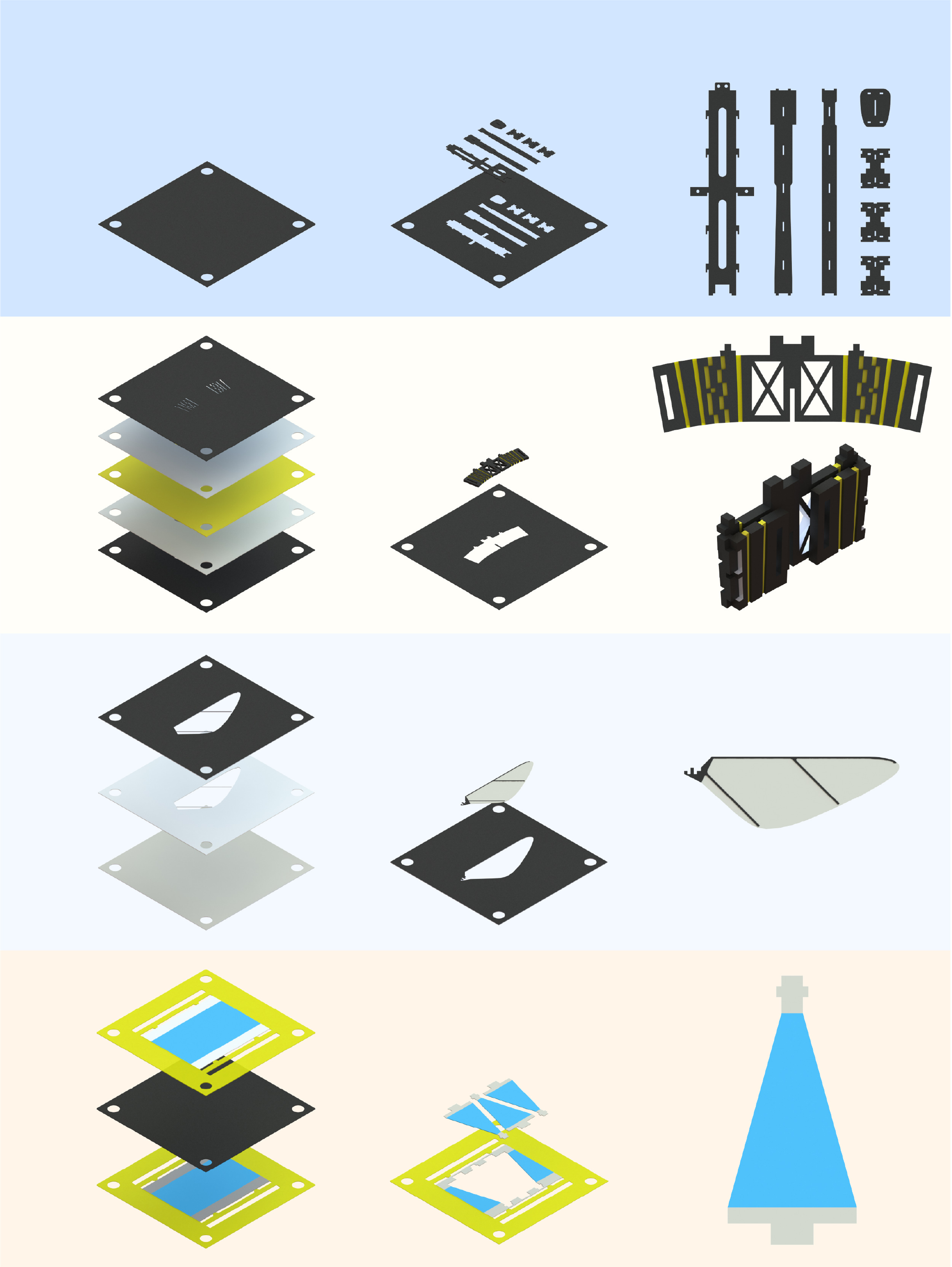}
\put (1,77)    {(a)}
\put (1,52)    {(b)}
\put (1,27)    {(c)}
\put (1,2)     {(d)}
\put (12,96)   {Step 1}
\put (35.5,96) {Step 2}
\put (59,96)   {Step 3}
\end{overpic}
\vspace{-1ex}
\caption{Fabrication processes of the components of the RoboBee and Bee\textsuperscript{+}: (a)~airframe, (b)~transmission mechanism, (c)~wing, (d)~actuators. Where the parts differ between the two FWMAVs, the version corresponding to the RoboBee is depicted. Step\,1 is the creation of the stack, which may contain layers that need to be cured. Step\,2 is the release cutting of the parts and Step\,3 shows the parts ready for assembly. The RoboBee transmission mechanism is the only part that must be folded into position before it can be used. \label{FIG02}}
\vspace{-2ex}
\end{figure}

\vspace{-0.5ex}
In the fabrication of both robots, we employ the SCM method presented in~\cite{wood2003microrobotics}. Accordingly, each component is made individually, then assembled manually under a microscope. The fabrication of each component of the robots in Fig.\,\ref{FIG01} follows the general procedure shown in Fig.\,\ref{FIG02}. First, the layers that comprise a composite monolithic part are micromachined using a precision \textit{diode-pumped solid-state} (DPSS) laser (Photonics Industries, DCH-355-3) with a beam diameter of 10\,$\mu$m; then, these featured pieces are assembled into a pin-aligned stack between two aluminum plates. Depending on the piece, a stack may contain uncured sheets of carbon fiber pre-preg or a sheet of adhesive that with the application of high temperature and pressure bond the entire assembly together. The resulting featured 2D monolithic pieces are then cut to release the foldable and assemblable parts used to create 3D micro-components, including the structures, mechanisms and actuators that compose the microrobots. Note that despite the fact that the actuation mechanisms of the two robots in Fig.\,\ref{FIG01} are markedly different, the other components are very similar and the assembly processes are almost identical.   

\vspace{-1ex}
\subsection{Common Elements}
\label{SEC03A}
\vspace{-0.5ex}
First, we describe the fabrication procedures for making parts that are either identical or very similar in both the RoboBee and Bee\textsuperscript{+}.
\subsubsection{Airframes}
An airframe consists of pieces made from either cured carbon fiber or fiberglass composite FR4. To make one sheet of cured carbon fiber, multiple layers of unidirectional carbon fiber pre-preg (Teijin Carbon Tenax\textsuperscript{\textregistered} Prepreg) are layered and placed into an automatic hydraulic press, where they receive a high-pressure treatment at 130\,$^\circ$C. The number and direction of these layers are determined by the final use of the cured carbon fiber. The carbon fiber laminas employed to make airframes are composed of four layers oriented perpendicularly with respect to each other in order to preset adequate strength and stiffness. The cured carbon fiber and FR4 pieces are laser-cut featured, laser-cut released and prepared for assembly as shown in~Fig.\,\ref{FIG02}(a), according to Step\,1~to~\,3.
\subsubsection{Transmissions and Hinges} 
These components are made from cured carbon fiber, utilized as the main structural material, and Kapton, utilized as the main elastic material. In the fabrication process, sheets of adhesive (Dupont Pyralux FR) are employed to bond the layers of Kapton and carbon fiber through the application of pressure and heat using an automatic hydraulic press (117\,psi at 180\,$^\circ$C for an hour). The RoboBee's two transmissions are made from a layer of Kapton inserted between two layers of carbon fiber and is folded manually to obtain the final configuration shown in~Fig.\,\ref{FIG02}(b). These transmissions are symmetric and map the approximately linear actuator outputs ($\delta$ in Fig.\,\ref{FIG01}(b)) to the flapping motions of the corresponding wings ($\varphi$ in Fig.\,\ref{FIG01}(b)). The Bee\textsuperscript{+} prototype has one independent transmission per wing, whose design, based on that presented in~\cite{ma2015mechanical}, does not require a manual folding step. The hinges for both robots are identically made using the method in~\cite{ma2015mechanical}, which consists of inserting a Kapton layer between two featured layers of carbon fiber that are then cured.  
\subsubsection{Wings} 
The wings for both robots are composed of a structural frame and spars made from cured carbon fiber pre-preg, and a membrane made from polyester film (see Fig.\,\ref{FIG02}(c)). Sheets of adhesive (Dupont Pyralux FR) are employed to bond the pieces of Mylar and carbon fiber through the application of pressure and heat (117\,psi at 180\,$^\circ$C for an hour). Each structural lamina for a wing is made by curing under pressure and heat (45\,psi at 130\,$^\circ$C for two hours) two layers of unidirectional carbon fiber pre-preg, aligned along the direction of the wing's leading edge, and an additional layer of pre-preg oriented at angle of 45$^\circ$ with respect to the wing's leading edge to provide adequate strength and elasticity to the spars.

\vspace{-1ex}
\subsection{Fabrication\,of\,Bimorph\,Actuators\,for\,RoboBee\,Prototypes}
\label{SEC03B}
\vspace{-0.5ex}
The stacks employed to make bimorph actuators are composed of a layer of alumina and a layer of piezoelectric PZT (lead zirconate titanate) physically aligned and constrained within an FR4 frame, and a central layer of unidirectional carbon fiber pre-preg (Torayca\textsuperscript{\textregistered} M46J). Once the fabrication of an actuator is completed, the piece of carbon fiber has a structural function but also serves as the central electrode in the parallel connection of the actuator according to the configuration in~\cite{wood2007design}; chemically-inert non-conductive pieces of alumina are employed to connect the actuator with the airframe (base in Fig.\,\ref{FIG02}(d)) and the transmission mechanism (tip in Fig.\,\ref{FIG02}(d)); and the laminas of PZT serve as the piezoelectric \textit{active layers} that bend the biomorph actuator. To bond the pieces together and cure the carbon fiber pre-preg, a pressure of 15\,psi is applied to the stack using a weight, inside an oven at 180\,$^\circ$C for two hours. A final two-sided (from the top and bottom surfaces) laser cut procedure is executed to release several 25-mg bimorph actuators from each stack.

\vspace{-1ex}
\subsection{Fabrication\,of\,Unimorph\,Actuators\,for\,Bee\textsuperscript{+}\,Prototypes}
\label{SEC03C}
\vspace{-0.5ex}
The most important components of a Bee\textsuperscript{+} prototype are the two pairs of twinned unimorph actuators. The corresponding fabrication process is a modification of the method employed to make bimorph actuators. The main difference is that the fabrication stack contains only one PZT layer; therefore, each resulting actuator has only one piezoelectric active layer, instead of two. Consistently, the final release requires only one laser cut from the top surface of the stack. This simplified procedure not only reduces the releasing time by half, but also improves the yield of usable actuators per stack.

\vspace{-1.0ex}
\subsection{Assembly of the Robots}
\label{SEC03D}
\vspace{-0.5ex}
The assembly of all the components into one functional system is the most crucial step in the fabrication of the microrobots. The processes for the two robotic insects considered here are very similar. The procedure begins with three separate subassemblies and ends by joining the resulting three subcomponents. Thus, we first assemble the airframe utilizing tab-and-slot features to position and orient the pieces relative to each other, which are then permanently joined and fixed with \textit{cyanoacrylate} (CA) glue. Secondly, the wings are attached with CA glue to their respective hinges using toothed matching features. Thirdly, the actuators are attached to their respective transmissions. In the cases of both microrobots, tabs on the tips of the actuators are glued to matching slots in the transmissions. To continue, the actuator-transmission subassemblies are glued to the base of the airframe. We utilize orthogonal contact surfaces between the actuators and the base as constraints, in order to enforce precision. Then, the \textit{ground linkages} of the transmissions are affixed to the airframe using CA glue. Finally, the wing-hinge subassemblies are attached to the transmissions; as their assembly relationships are only loosely constrained, manual adjustments can be made to compensate for assembly errors incurred in prior steps. In some cases, protective spars and legs, micromachined from cured carbon fiber, are added to statically stabilize the robot on flat surfaces and prevent damage to the body in case of a crash.

\vspace{-1ex}
\section{Aerodynamics of the RoboBee and Bee\textsuperscript{+}}
\label{SEC04}
\vspace{-0.5ex}
\label{aero}
The wings of both the RoboBee and Bee\textsuperscript{+} are designed with the same geometric parameters and to operate with similar kinematics at low Reynolds numbers. If the \textit{clap-and-fling} mechanism is not employed for control, the aerodynamics of both platforms can be analyzed with the methods and tools in~\cite{whitney2010aeromechanics, chang2016dynamics,chang2018time,yang2019bee}. From the analyses in~\cite{yang2019bee}, it follows that Bee\textsuperscript{+} has the potential to outperform the RoboBee in many aspects when factors such as the number of actuation inputs, yaw damping, steering capabilities and wing-loading are considered. To explain the need for active torque regulation in both platforms, we provide a simplified qualitative analysis of the process of aerodynamic torque generation during flight. 

\begin{figure}[t!]
\centering
\subfloat[(a)~RoboBee and frames.]{
\includegraphics{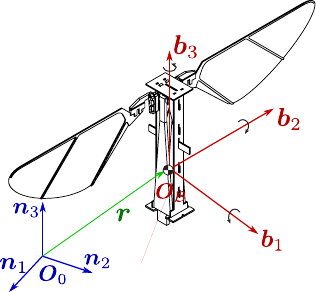}}
\hspace{0.2cm}
\subfloat[(b)~Bee\textsuperscript{+} and frames.]{
\includegraphics{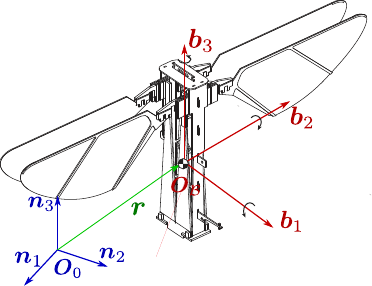}}
\caption{Frames of reference employed to model the two insect-scale FWMAVs used in this work. In~(a), the frames of reference of the RoboBee are indicated. In (b), the frames of reference of Bee\textsuperscript{+} are indicated. For both robots, $\left\{\bs{n}_1,\bs{n}_2,\bs{n}_3\right\}$ is the inertial frame and $\left\{\bs{b}_1,\bs{b}_2,\bs{b}_3\right\}$ represents the body-fixed frame, whose origin coincides with the center of mass of the robot. \label{FIG03}}
\vspace{-2ex}
\end{figure}

For a flying robot, as discussed in~\cite{wood2008first,perez2011first,perez2015model}, the capability to liftoff is the first requirement to be achieved; therefore, for the sake of analysis, we consider this simple case. For a perfectly-fabricated RoboBee, in the absence of external disturbances while flying in open loop, if each wing is flapped symmetrically with respect to the $\boldsymbol{b}_2\boldsymbol{b}_3$ plane defined in~Fig.\,\ref{FIG03}(a), the cycle-averaged force is expected to lie inside the $\boldsymbol{b}_2\boldsymbol{b}_3$ plane defined in~Fig.\,\ref{FIG03}(a). Consistently, from the cycle-averaged perspective, no pitch motion about the $\boldsymbol{b}_2$ axis is induced. From the instantaneous viewpoint, however, when a wing flaps away from the $\boldsymbol{b}_2\boldsymbol{b}_3$ plane, an instantaneous pitch torque is generated and the robot's body might deviate from its initial zero pitch attitude; since the robot is moving upward, the down-wash airflow generates a nonzero cycle-averaged downward force that might induce the body to pitch. If this condition occurs, the pitch motion is dominated by a destabilizing mode and the response of the entire system becomes unstable. 

For a perfectly-fabricated Bee\textsuperscript{+}, in the absence of external disturbances while flying in open loop, if the four wings are flapped according to a perfectly synchronized and symmetric sinusoidal pattern  with respect to the planes $\boldsymbol{b}_1\boldsymbol{b}_3$ and $\boldsymbol{b}_2\boldsymbol{b}_3$ in Fig.\,\ref{FIG03}(b), the instantaneous pitch torque generated by each wing is canceled by the pitch torque generated by its twin wing. Therefore, body-pitch motions can not be induced by the instantaneous flapping motion of the wings. However, if a small pitch angle already exists, due to the existence of fabrication errors or an initial condition, a nonzero cycle-averaged force due to down-wash airflow appears, which might induce the pitch response to become unstable. Because the mean position of each wing lies outside the $\boldsymbol{b}_2\boldsymbol{b}_3$ plane, Bee\textsuperscript{+} can pitch significantly faster than the RoboBee, due to a longer torque arm. For both robots in experimental tests, disturbances induced by fabrication errors, external disturbances and horizontal drifting are impossible to avoid; therefore, active pitch torque regulation is necessary even for simple tasks such as vertical take-off. Similarly, torque regulation is required to compensate for undesired roll and yaw rotations and to actively steer the robot during flight.

Control methods to regulate the pitch and roll motions of artificial insects have been already introduced and discussed in~\cite{finio2012open, ma2013controlled, perez2015model,fuller2019four}. However, as explained in~\cite{gravish2016anomalous,fuller2019four,yang2019bee}, effective methods to regulate the yaw motion of FWMAVs have not yet been developed. To achieve yaw steering capabilities, three yaw-torque-generation strategies are theoretically implementable~\cite{yang2019bee}. Namely, the \textit{split-cycle} scheme, which generates upstroke and downstroke flapping motions with significantly different frequencies, and thus might require a wide actuation bandwidth; the \textit{asymmetric angle of attack} scheme, which employs specialized actuation mechanisms to actively adjust the upstroke and downstroke angle of attack; and the \textit{inclined stroke-plane} scheme, which tilts the wing-stroke-plane to project a nonzero force component onto the steering plane ($\boldsymbol{b}_1\boldsymbol{b}_2$) and pairs the diagonal wings of the robot to generate a nonzero yaw torque. For the RoboBee, the implementation of the split-cycle scheme is feasible; as discussed in~\cite{gravish2016anomalous,fuller2019four}, however, this scheme is experimentally not very effective. In contrast, compared to the RoboBee, Bee\textsuperscript{+} has a theoretically wider bandwidth due to its smaller stroke amplitude and, more importantly, the implementation of the inclined stroke-plane method is feasible from both the theoretical and experimental perspectives. Thus, we believe that the Bee\textsuperscript{+} configuration is very promising to achieve effective controlled yaw steering.  
\captionsetup[subfigure]{labelformat=empty}

\vspace{-1ex}
\section{Flight Controller Design}
\label{SEC05}
\vspace{-0.5ex}
\subsection{System Dynamics}
\label{SEC05A}
\vspace{-0.5ex}
\captionsetup[subfigure]{labelformat=empty}
Both the Robobee and Bee\textsuperscript{+} can be modeled as a single rigid body, as done in~\cite{Ying2016ICRA,chen2017lyapunov} to describe the dynamics of a quadrotor. The frames of reference of both robots are shown in Fig.\,\ref{FIG03}; here, $\left\{\bs{n}_1,\bs{n}_2,\bs{n}_3\right\}$ represents the inertial frame and $\left\{\bs{b}_1,\bs{b}_2,\bs{b}_3\right\}$ represents the body-fixed frame with the origin located at the center of mass of the respective robot. For both robots, the direction of the total thrust force is assumed to be aligned with the $\bs{b}_3$ axis and the number of actuators is less than the total number of degrees of freedom of the system; therefore, both robots are underactuated thrust-propelled systems. Accordingly, the globally defined nonlinear dynamics of both robots is described by
\begin{align}
m \ddot{\bs{r}}
&= -mg \bs{n}_3 + f \bs{b}_3, \label{EQ01} \\
\dot{\bs{\qbar}}& = \frac{1}{2}\bs{\qbar}\ast\bs{\pbar}, \label{EQ02} \\
\bs{J} \dot{\bs{\omega}} &= -\bs{\omega} \times \bs{J} \bs {\omega} 
+ \bs{\tau}, \label{EQ03}  
\end{align} 
where $m$ is the total mass of the robot; $\bs{r} = \left[r_{1}\;r_{2}\;r_{3}\right]^T$ indicates the displacement of the robot's center of mass with respect to the origin of the inertial frame; and $f$ is the magnitude of the total thrust force generated by the flapping wings. Quaternion $\bs{\qbar}$ is employed to describe the attitude of the robot with respect to the inertial frame; quaternion $\bs{\pbar} = \left[0 \;\; \bs{\omega}^T\right]^T$; and the symbol $\ast$ represents the standard multiplication of quaternions. $\bs{J}$ denotes the moment of inertia of the robot; $\bs{\omega}$ is the flyer's angular velocity expressed in the body frame with respect to the inertial frame; and $\bs{\tau}=\left[\tau_1\;\tau_2\;\tau_3\right]^T$ is the torque generated by the flapping wings. 

In (\ref{EQ01})--(\ref{EQ03}), several assumptions are made. Most notably, the direction of the thrust force is assumed to be aligned with $\bs{b}_3$; the projection of the total aerodynamic force generated by the flapping wings onto the steering plane $\boldsymbol{b}_1\boldsymbol{b}_2$ is assumed to be negligible in one flapping cycle, which implies that $f\bs{b}_3$ is the only external actuation force; all the aerodynamic disturbances affecting the system are considered to be negligible; and the gyroscopic effect between the wing flapping and body rotation is ignored. Furthermore, the model ignores the secondary effects produced by the yaw-torque-generation mechanism employed by the Bee\textsuperscript{+}, described in Section\,\ref{SEC04} and \cite{yang2019bee}, as the inclined angle between the stroke plane and steering plane is not modeled.  
\vspace{-1ex}
\subsection{Actuator Command Generation}
\label{SEC05B}
\vspace{-0.5ex}
The piezoelectric actuators of both robots are driven by high-frequency (100\,Hz) sinusoidal signals. For both prototypes, the determination of the mapping from the desired thrust force and torque to the parameters that define the sinusoidal signals of actuation (force\,\&\,torque/actuation mapping for short) is essential to achieve desired closed-loop performances. The force\,\&\,torque/actuation mapping of the RoboBee prototype is given by
\begin{align}
\begin{bmatrix}
f \\ \tau_1 \\ \tau_2 \\ \tau_3 
\end{bmatrix} = 
\begin{bmatrix}
k_{\ts{amp}} & 0 & 0 & 0 \\
0 & k_{\ts{roll}} & 0 & 0 \\
0 & 0 & k_{\ts{pitch}} & 0 \\
0 & 0 & 0 & k_{\ts{yaw}}  
\end{bmatrix}
\begin{bmatrix}
\theta_{\ts{amp}} \\ \theta_{\ts{roll}} \\ \theta_{\ts{pitch}} \\ \theta_{\ts{yaw}}
\end{bmatrix}, 
\label{EQ04}
\end{align}
where $f$ and $\left[ \tau_1 \; \tau_2 \; \tau_3 \right]^T$ are the magnitude of the thrust force and torque required by the controller of the flyer, respectively; $\theta_{\ts{amp}}$ is the mean flapping amplitude of both wings; $\theta_{\ts{roll}}$ is the differential flapping angle between the two wings; $\theta_{\ts{pitch}}$ is the shift of the mean flapping angle; $\theta_{\ts{yaw}}$ is the proportion of the second-harmonic sinusoidal signal, according to the split-cycle method described in~\cite{ma2013controlled}; and $k_{\ts{amp}}$, $k_{\ts{roll}}$, $k_{\ts{pitch}}$ and $k_{\ts{yaw}}$ denote the coefficients of the mapping. The computation of the inverse mapping specified by (\ref{EQ04}) is straightforward. 
    
Now, we focus on determining the actuation mapping of the Bee\textsuperscript{+} robot. Simplifying the model introduced in~\cite{yang2019bee}, the thrust force generated by the $i$th flapping wing is estimated as $f_i = k_f v_i$, for $i = 1,\cdots,4$, where $v_i$ is the magnitude of the sinusoidal command signal generated by the $i$th unimorph actuator; and $k_f$ is the coefficient that relates the thrust force with the command magnitude. In theory, yaw torques in the steering plane can be generated by employing either the split-cycle method or the inclined stroke-plane scheme described in Section\,\ref{SEC04}. The projected component of the aerodynamic force on the steering plane, produced by the $i$th wing, is modeled as ${f_{\ts{s}}}_i = k_{\ts{s}} v_i$, for $i = 1,\cdots,4$, where $k_{\ts{s}}$ is the coefficient that relates the force with the amplitude of the command. Thereby, the force\,\&\,torque/actuation mapping of the Bee\textsuperscript{+} prototype is given by
\begin{align}
\begin{bmatrix}
f \\ \tau_1 \\ \tau_2 \\ \tau_3 
\end{bmatrix} = 
\begin{bmatrix}
k_f & k_f & k_f & k_f \\
-k_f d_1 & -k_f d_1 & ~k_f d_1 & ~k_f d_1\\
~k_f d_2 & -k_f d_2 & ~k_f d_2 & -k_f d_2 \\
~k_{\ts{s}} d_3 & -k_{\ts{s}} d_3 & -k_{\ts{s}} d_3 & ~k_{\ts{s}} d_3
\end{bmatrix}
\begin{bmatrix}
v_1\\ v_2 \\ v_3 \\ v_4
\end{bmatrix}, \label{EQ05}
\end{align}
where $d_i$ is the lever arm associated with the torque $\tau_i$ required by the controller of the system. It is straightforward to see that the mapping in (\ref{EQ05}) is similar to the one that relates the rotors' speeds with the control force and torques in the quadrotor case presented in~\cite{Ying2016ICRA}. Consistently, the four-winged design of Bee\textsuperscript{+} increases the control capabilities of the system compared to those exhibited by the two-winged design in~\cite{ma2013controlled} as the thrust force and the control torques are generated by four wings rather than two. Also, note that the split-cycle-based yaw-torque-generation method introduced in~\cite{ma2013controlled} necessitates a wide bandwidth of actuation to apply the required second-harmonic sinusoidal signal, which is not experimentally feasible due to limitations of piezoelectric actuators; the four-winged design eliminates this drawback by using the projected components of the thrust forces to generate yaw torque, if the inclined stroke-plane scheme is implemented. In summary, taking the inverse of (\ref{EQ05}) yields the mapping that relates the thrust force and torques with the actuator commands as
\begin{align}
\begin{bmatrix}
v_1 \\ v_2 \\ v_3 \\ v_4
\end{bmatrix} =
\begin{bmatrix}
\frac{1}{4k_f} & -\frac{1}{4 d_1 k_f} & \frac{1}{4 d_2 k_f} &\frac{1}{4 d_3 k_{\ts{s}}} \\
\frac{1}{4k_f} & -\frac{1}{4 d_1 k_f} & -\frac{1}{4 d_2 k_f} & -\frac{1}{4 d_3 k_{\ts{s}}} \\
\frac{1}{4k_f} & \frac{1}{4 d_1 k_f} & \frac{1}{4 d_2 k_f} & -\frac{1}{4 d_3 k_{\ts{s}}} \\
\frac{1}{4k_f} & \frac{1}{4 d_1 k_f} & -\frac{1}{4 d_2 k_f} & \frac{1}{4 d_3 k_{\ts{s}}}
\end{bmatrix}
\begin{bmatrix}
f \\ \tau_1 \\ \tau_2 \\ \tau_3 
\end{bmatrix}. 
\label{EQ06}
\end{align}

\vspace{-1ex}
\subsection{Attitude Control}
\label{SEC05C}
\vspace{-0.5ex}
The desired attitude kinematics in quaternion form is given~by
\begin{align}
\bs{\dot{\qbar}}_{\ts{d}} = \frac{1}{2}\bs{\qbar}_{\ts{d}}\ast\bs{\pbar}_{\ts{d}}, 
\label{EQ07}
\end{align}
where $\bs{\qbar}_{\ts{d}}$ is the quaternion that represents the desired attitude and $\bs{\pbar}_{\ts{d}} = \left[0\;\;\bs{\hat{\omega}}_{\ts{d}}^T\right]^T$, in which $\bs{\hat{\omega}}_{\ts{d}}$ denotes the desired angular velocity expressed in the desired frame $\bs{\qbar}_{\ts{d}}$. It follows that the attitude error between $\bs{\qbar}$ and $\bs{\qbar}_{\ts{d}}$ can be described by the quaternion 
\begin{align}    
\begin{bmatrix}
m_\ts{e}  \\
\bs{n}_{\ts{e}}
\end{bmatrix}
=
\bs{\qbar}_{\ts{e}} = \bs{\qbar}_{\ts{d}}^{-1}\ast\bs{\qbar}. 
\label{EQ08}
\end{align}
Accordingly, we specify the attitude control torque to be
\begin{align}
\bs{\tau} = -\bs{K}_1\ts{sgn}(m_{\ts{e}})\bs{n}_{\ts{e}} - \bs{K}_2(\bs{\omega} - \bs{\omega}_{\ts{d}}) + \bs{\tau}_{\ts{d}}, \label{EQ09}
\end{align}
where $\bs{K}_1$ and $\bs{K}_2$ are positive definite diagonal gain matrices; $\ts{sgn}(\cdot)$ represents the sign function; $\bs{\omega}_{\ts{d}}$ denotes the desired angular velocity that has the exact same components as $\bs{\hat{\omega}}_{\ts{d}}$, but is expressed in the body frame $\bs{\qbar}$ instead of the desired frame ${\bs{\qbar}_{\ts{d}}}$; and $\bs{\tau}_{\ts{d}}$ represents the torque required to excite the desired angular velocity dynamics. The axis of the rotation from ${\bs{\qbar}}$ to ${\bs{\qbar}_{\ts{d}}}$ is denoted by the unit vector $\bs{a}_{\ts{e}}$ and the associated rotation angle is defined to be $\Theta_{\ts{e}}$, with $0 \leqslant \Theta_{\ts{e}} <\pi$. Then, the term $-\ts{sgn}(m_{\ts{e}})\bs{n}_{\ts{e}}$ is geometrically equivalent to $\sin{(\frac{1}{2}\Theta_{\ts{e}})}\bs{a}_{\ts{e}}$. Note that the multiplication of the term $\ts{sgn}(m_{\ts{e}})$ is employed to remove the ambiguity of the quaternion representation as ${\bs{\qbar}_{\ts{e}}}$ and ${-\bs{\qbar}_{\ts{e}}}$ indicate the same rotation result.

\vspace{-1ex}
\subsection{Position Control}
\label{SEC05D}
\vspace{-0.5ex}
From the control perspective, the dynamics of both the RoboBee and Bee\textsuperscript{+} are underactuated with the direction of the thrust force aligned with the $\bs{b}_3$ axis. Consistently, the position control of these systems requires the specification of the magnitude and orientation of the thrust force, as done in the cases of other flapping-wing robots and the quadrotors in~\cite{ma2013controlled,roberts2011adaptive}. Here, we propose a position controller that is comprised of two sub-algorithms. The first sub-scheme generates the magnitude of the thrust force, $f$; the second sub-scheme generates the desired attitude. In specific, $f$ is computed as 
\begin{align}
f &= \bs{f}_{\ts{a}}^T\bs{b}_3, \label{EQ10} \\
\begin{split}
\bs{f}_{\ts{a}} &= -\bs{K}_{\ts{p}}(\bs{r} - \bs{r}_{\ts{d}}) - \bs{K}_{\ts{d}}(\bs{\dot{r}} - \bs{\dot{r}}_{\ts{d}})  \\
&\quad - \bs{K}_{\ts{i}}\int(\bs{r} - \bs{r}_{\ts{d}}) dt + mg\bs{n}_{3} + m\bs{\ddot{r}}_{\ts{d}}, 
\label{EQ11}
\end{split}
\end{align}
where $\bs{K}_{\ts{p}}$, $\bs{K}_{\ts{d}}$ and $\bs{K}_{\ts{i}}$ are positive definite diagonal gain matrices; and $\bs{r}_{\ts{d}}$ is the desired position of the robot's center of mass. Due to the underactuation of the system, the desired thrust force $\bs{f}_{\ts{a}}$ can not be directly generated by the microrobots and must be coordinately produced with the attitude of the robot; in particular, the direction of the $\bs{b}_3$ axis. Accordingly, the desired attitude is derived from the desired thrust force $\bs{f}_{\ts{a}}$ and the desired yaw angle $\psi_{\ts{d}}$ as
\begin{align}
{\bs{b}_{3}}_{\ts{d}} = &  \frac{\bs{f}_{\ts{a}}}{\left|\bs{f}_{\ts{a}}\right|_2}, \label{EQ12} \\
{\bs{b}_{1}}_{\ts{d}} = &  \frac{\left[-\sin{\psi_{\ts{d}}}\; \cos{\psi_{\ts{d}}}\; 0\right]^T\times {\bs{b}_{3}}_{\ts{d}}}{\left|  \left[-\sin{\psi_{\ts{d}}}\; \cos{\psi_{\ts{d}}}\; 0\right]^T\times {\bs{b}_{3}}_{\ts{d}}\right|_2}, \label{EQ13} \\
{\bs{b}_{2}}_{\ts{d}} = & {\bs{b}_{3}}_{\ts{d}}\times {\bs{b}_{1}}_{\ts{d}}, \label{EQ14}
\end{align}
where ${\bs{b}_1}_{\ts{d}},{\bs{b}_2}_{\ts{d}},{\bs{b}_3}_{\ts{d}}$ are the desired axes of the body frame expressed in the inertial frame. Thus, directly from (\ref{EQ12})--(\ref{EQ14}), the rotation matrix $\bs{S}_{\ts{d}} = \left[{\bs{b}_1}_{\ts{d}}\; {\bs{b}_2}_{\ts{d}}\; {\bs{b}_3}_{\ts{d}}\right]$ that describes the desired attitude of the robot is arranged. From this matrix, the desired attitude quaternion ${\bs{\qbar}_{\ts{d}}}$ and the corresponding kinematics specified by (\ref{EQ07}) are computed.

\vspace{-1ex}
\section{Experimental Results}
\label{SEC06}
\vspace{-0.5ex}
\subsection{Experimental Setup}
\label{SEC06A}
\vspace{-0.5ex}
\begin{figure}[t!]
\vspace{2ex}
\begin{center}
\includegraphics[width=0.96\linewidth]{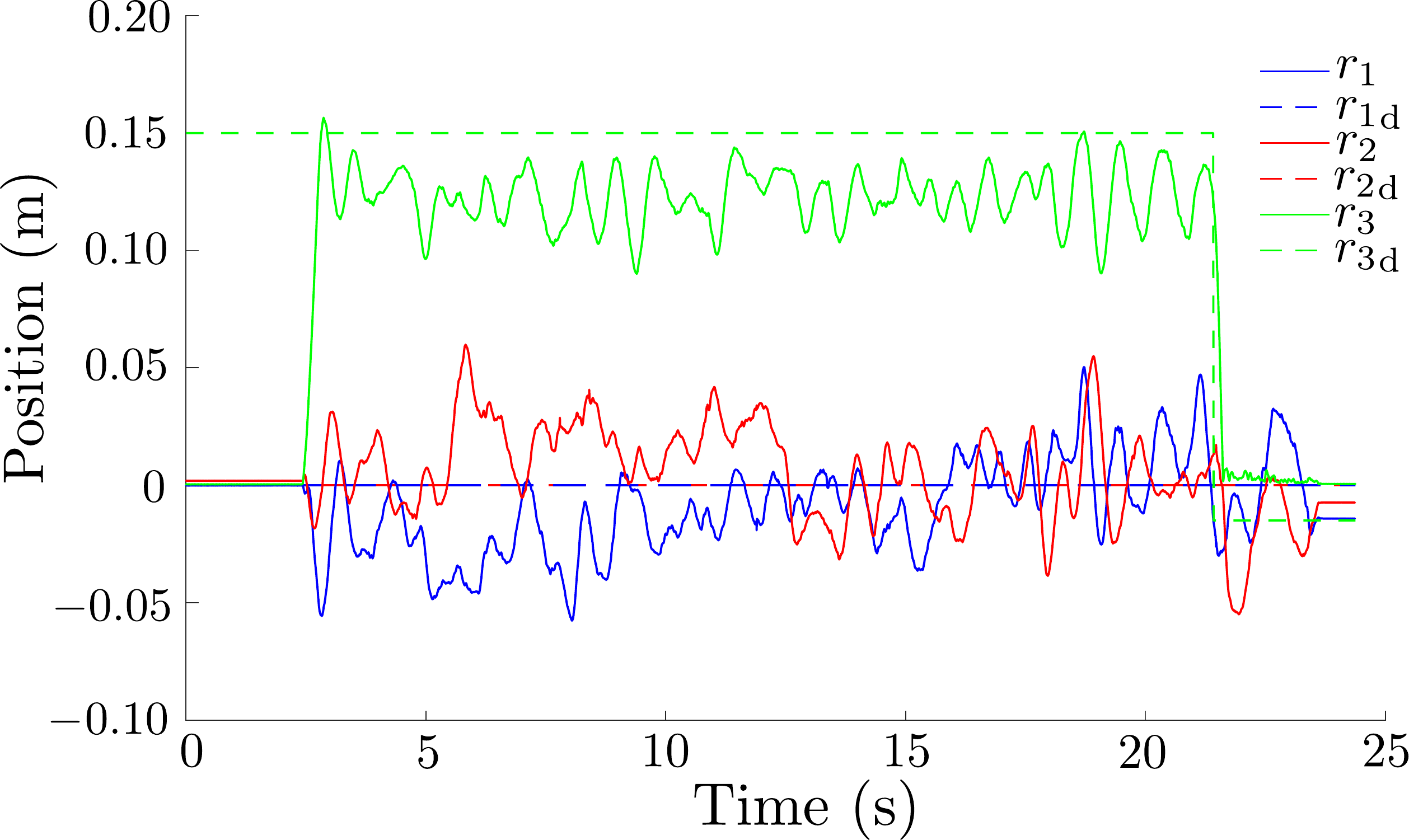}
\end{center}
\vspace{-1ex}
\caption{Position control of the RoboBee prototype. The dashed lines represent the reference position signals and the solid lines represent the measured positions. In this case, the robot is commanded to hover at a desired position. The entire experiment lasts for almost \SI{20}{\second}, which verifies the stability robustness and performance consistency of the attitude and position controllers proposed in Sections~\ref{SEC05C}~and~\ref{SEC05D}. \label{FIG04}}
\vspace{1ex}
\begin{center}
\includegraphics[width=0.96\linewidth]{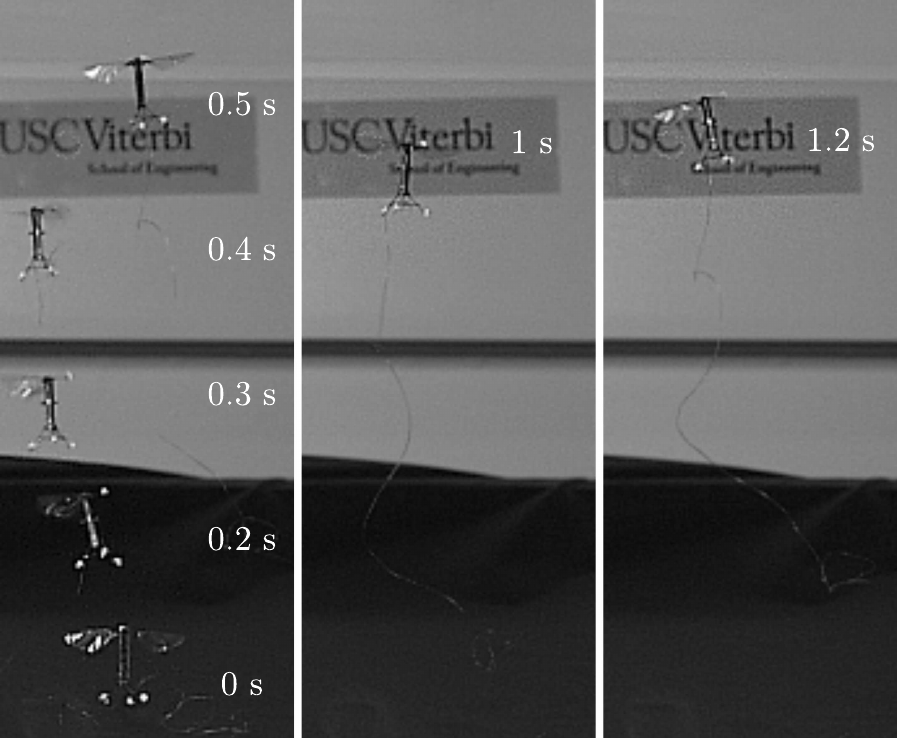}
\end{center}
\vspace{-1ex}
\caption{Photographic sequence of the RoboBee prototype during a position flight control experiment. The robot is able to hover around the desired position for about \SI{20}{\second}. \label{FIG05}}
\vspace{-2ex}
\end{figure}
The experimental setup includes the insect-scale flapping-wing robots, piezo-actuator drivers (PiezoMaster VP7206), a \textit{Vicon motion capture} (VMC) system, and a ground computer, running in a Mathworks Simulink Real-Time hardware--software configuration, that is employed for signal processing and generating the real-time control signals. The control algorithms are run at the frequency of \SI{2}{\kilo\hertz} and the VMC system measures the position and attitude states at the frequency of \SI{500}{\hertz}. As the angular velocity of the robot can not be directly measured with the VMC system, this is estimated as
 \begin{align}
\begin{bmatrix}
0 \\
\bs{\omega} 
\end{bmatrix}
= 2\bs{\qbar}^{-1}\ast\left[\frac{\lambda s}{s + \lambda}\right]\bs{\qbar}, 
\label{EQ15}
\end{align}
where the bracket $\left[\cdot\right]$ represents a low-pass filter that operates on the signal ${\bs{\qbar}}$; $\lambda$ is a constant parameter; and $s$ is the complex variable of the Laplace transform. A similar low-pass derivative filter is employed to operate on the position states in order to estimate the translational velocities. Note that the utilization of low-pass filters is necessary to estimate the velocities of the two robots because high-frequency flapping unavoidably induces high-frequency oscillations and, therefore, high-frequency noise. The overall controller design and real-time implementation approach is experimentally validated in multiple ways. For example, the open-loop trimming flight tests, required in the cases reported in~\cite{ma2013controlled,chirarattananon2014adaptive}, are not necessary to implement the controller introduced in this paper. This implies that, in the proposed approach, the fine tuning of the command signals to preset zero offset torques is not needed.

\vspace{-1ex}
\subsection{Experimental Results of the RoboBee}
\label{SEC06B}
\vspace{-0.5ex}
In this section, we present experimental results obtained with the implementation of the position and attitude controllers specified by (\ref{EQ09})--(\ref{EQ11}) on a RoboBee prototype that is enabled to hover at a desired position. During flight, the roll and pitch rotations are controlled to change the direction of the lift force to achieve position control. The yaw rotation is left in open-loop as the production of yaw torque is insufficient to generate effective regulation due to the limited bandwidth of the actuators~\cite{yang2019bee,fuller2019four}. The position in space of the robot during a controlled hovering flight experiment is shown in~Fig.\,\ref{FIG04}; the reference signals were plotted using dashed lines and the measurements were plotted using solid lines. In this case, the flyer first takes off, then hovers about the reference position in space, and finally lands back on the experimental table. The magnitudes of the instantaneous position errors along the $\bs{n}_1$ and $\bs{n}_2$ axes remain smaller than 7\,cm, and the instantaneous altitude error remains smaller than 6\,cm. Despite non-negligible position errors, this hovering experiment demonstrates that the proposed attitude and position regulation methods allow the robot to achieve the main control objective in a satisfactory manner as it can autonomously take off, hover and land without crashing or stalling.
 
The time-lapse of the flight test during the first 1.2\,s is shown in~Fig.\,\ref{FIG05}. The entire experiment lasts for almost \SI{20}{\second}, which demonstrates the stability robustness and performance consistency of the closed-loop system. The tracking errors seen in~Fig.\,\ref{FIG04} are partially caused by oscillations of the thrust-force direction that are produced by flapping-wing-induced oscillations of the robot's body. Another indirect source of tracking errors is the existence of unmodeled phenomena, such as the aerodynamic effects that are ignored in the mapping specified by (\ref{EQ04}). This issue can be addressed with high-order dynamic descriptions of the system, in combination with adaptive methods to estimate in real-time the unknown parameters, as done in~\cite{Ying2018IROS}. The complete experiment can be seen in the supplementary movie S1.mp4, also at
{\footnotesize \url{https://www.uscamsl.com/resources/ICAR2019/S1.mp4}}.
\begin{figure}[t!]
\vspace{2ex}
\begin{center}
\subfloat[(a) Reference and measured altitudes.]{
\includegraphics[width=\linewidth]{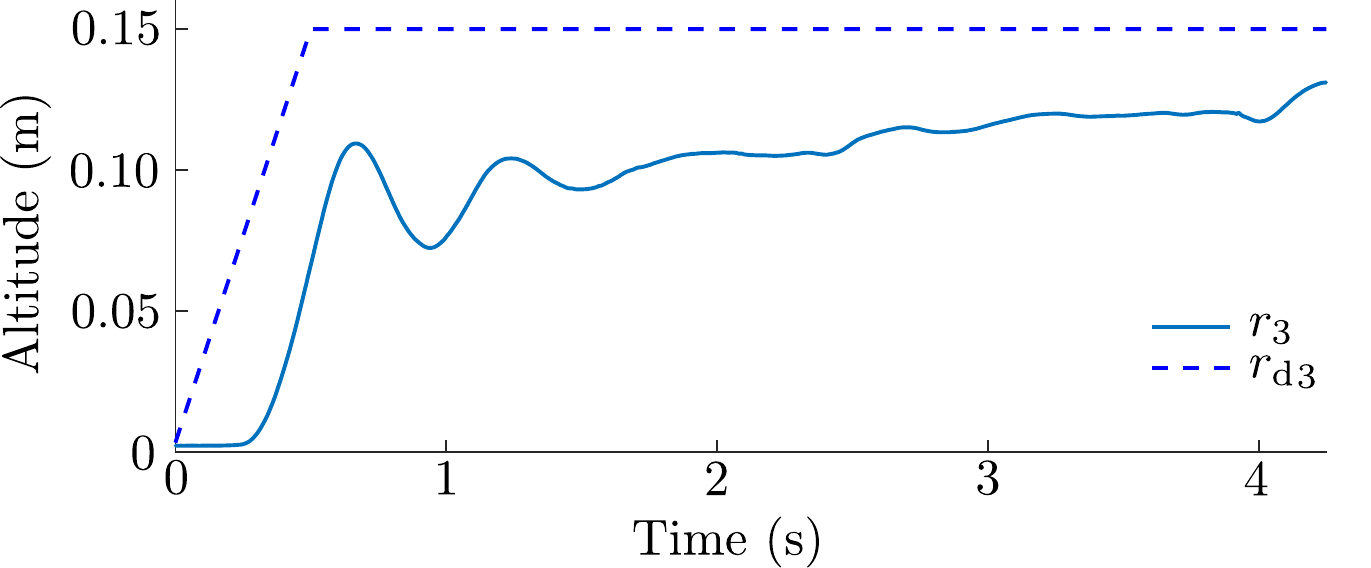}}
\hspace{10ex}
\subfloat[(b) Measured Euler roll and pitch angles.]{
\includegraphics[width=0.96\linewidth]{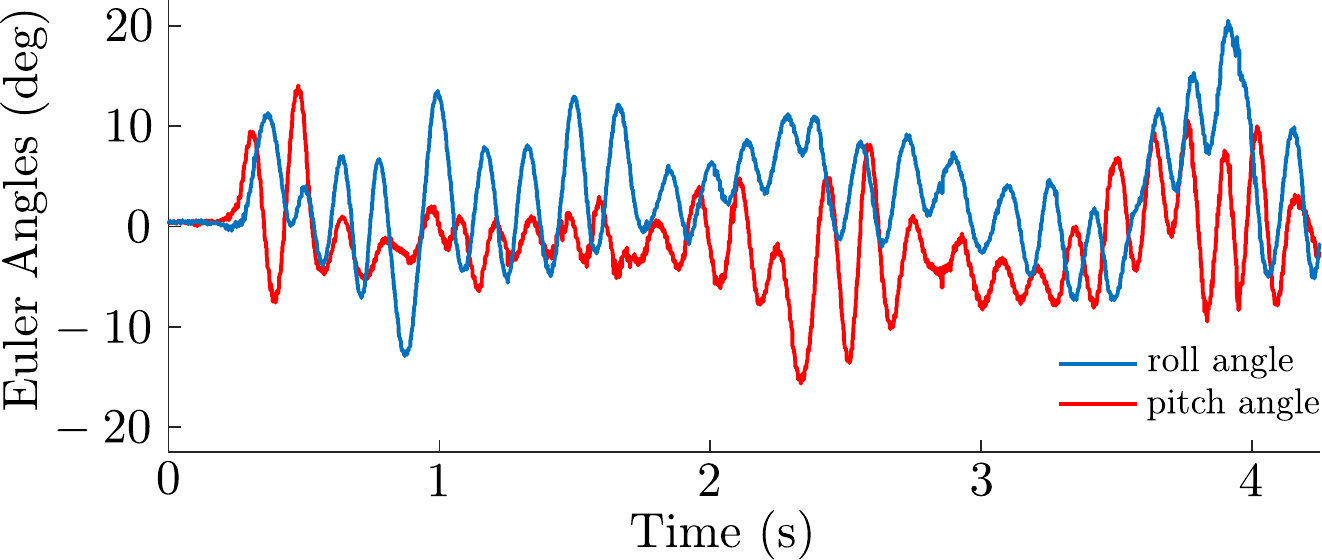}}
\caption{Altitude-and-attitude control experiment performed using the Bee\textsuperscript{+} prototype. (a)~This plot shows the altitude of the robot during flight. The dashed line represents the reference altitude and the solid line represents the measured signal. (b)~This plot shows the Euler roll and pitch angles of the robot during flight. This angular oscillation is partially caused by the periodic flapping of the wings and remains bounded between \ang{-10} and \ang{10}, which is experimentally acceptable. The experiment lasts for approximately \SI{5}{\second}; after this period of time, the robot leaves the specified safety volume and the power is turned automatically off. \label{FIG06}}
\end{center}	    
\vspace{1ex}
\begin{center}
\includegraphics[width=0.96\linewidth]{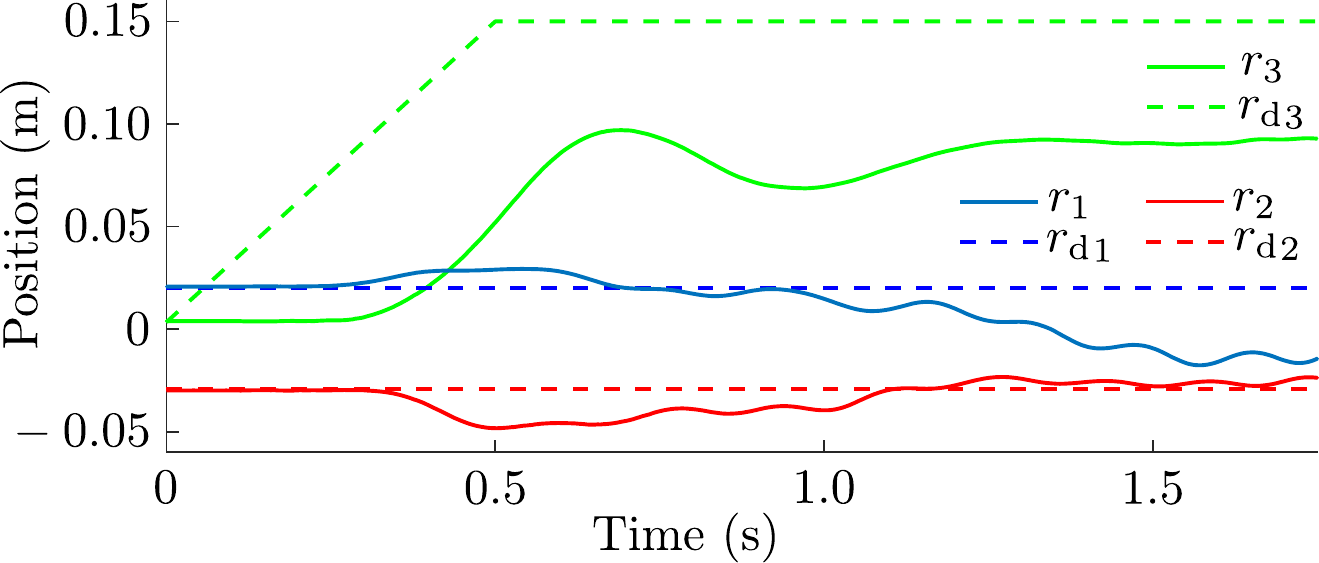}
\end{center}
\vspace{-1ex}	
\caption{Position control of the Bee\textsuperscript{+} prototype. The dashed lines represent the reference position signals and the solid lines represent the measured controlled positions. The experiment lasts for less than \SI{2}{\second} due to an increased oscillation of the pitch axis, which is caused by an insufficient total thrust force. \label{FIG07}}
\vspace{1ex}
\begin{center}
\includegraphics[width=0.96\linewidth]{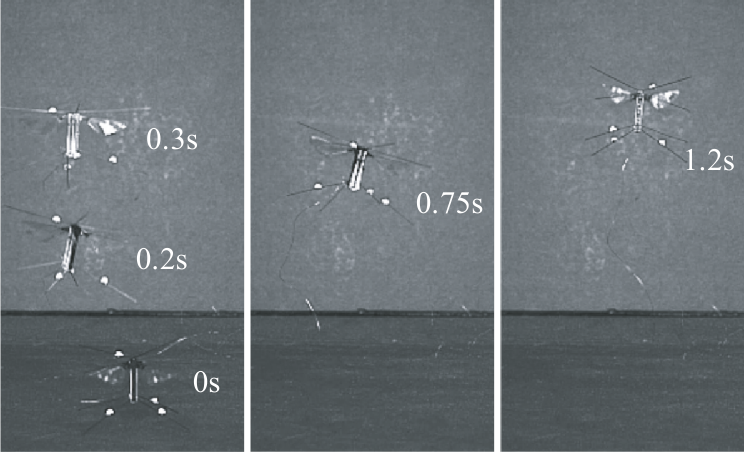}
\end{center}
\vspace{-1ex}
\caption{Photographic sequence of the Bee\textsuperscript{+} prototype during a position flight control experiment. From \SI{0}{\second} to \SI{0.3}{\second}, the robot takes off and rapidly reaches the reference position. It remains hovering around this point in space from \SI{0.3}{\second} to \SI{1.2}{\second}. \label{FIG08}}
\vspace{-2ex}
\end{figure}

\vspace{-1ex}
\subsection{Experimental Results of Bee\textsuperscript{+}}
\label{SEC06C}
\vspace{-0.5ex}
Experimental results obtained with a Bee\textsuperscript{+} prototype include data from both an altitude-and-attitude controlled flight and a position controlled flight. In the first case, the objective is to implement a controller to enable a Bee\textsuperscript{+} prototype to fly at a desired altitude while the direction of the thrust force remains perpendicular to the $\bs{n}_1\bs{n}_2$ plane. This goal is equivalent to regulating the altitude to a constant value, and the roll and pitch angles to zero. Yaw feedback control is not implemented in order to reduce the control burden on the flapping wings; therefore, the direction of the thrust force is unchanged by the yaw angle. A modified version of the position controller specified by (\ref{EQ10})--(\ref{EQ11}) is implemented to control the robot's altitude only, according to the method presented in~\cite{yang2019bee}. The results corresponding to an altitude-and-attitude controlled flight experiment are shown in~Fig.\,\ref{FIG06}. It is clear that the attitude controller given by (\ref{EQ09}) enables the thrust-force direction to be approximately perpendicular to the $\bs{n}_1\bs{n}_2$ plane with angular oscillations within the interval $\left[\SI{-10}{\degree}, \SI{10}{\degree}\right]$, which indicates that the magnitude of the lift force is not significantly reduced by attitude oscillations.

The last test discussed in the paper is the implementation of the position-and-attitude controller given by (\ref{EQ09})--(\ref{EQ11}) on the Bee\textsuperscript{+} robot to enable it to hover at a desired position. Just as in the altitude-and-attitude control case, the desired yaw angle is set to be the true yaw angle, which does not affect the position control. The corresponding experimental results are shown in~Fig.\,\ref{FIG07}, which compares the measured controlled position with the reference signals. Here, the Bee\textsuperscript{+} prototype is observed to approximately track the reference signals in the first second; then, the position error along the $\bs{n}_1$ axis gradually increases due to an increase in the pitch axis oscillation. We hypothesize that this phenomenon is caused by actuator saturation. We believe that this problem can be addressed by improving the robotic design to generate more lift and torque for position regulation and trajectory following flight. The time-lapse of the first 1.2\,s of the experiment is shown in Fig.\,\ref{FIG08}. The entire experiment can be seen in the supplementary movie S1.mp4, also available at {\footnotesize \url{https://www.uscamsl.com/resources/ICAR2019/S1.mp4}}.

\vspace{-1ex}
\section{Conclusions and Future Work}
\label{SEC07}
\vspace{-0.5ex}
We proposed a compact and robust multi-platform method for controlling FWMAVs in hovering flight, and its suitability was clearly demonstrated through experiments. The proposed controller design follows a unifying approach in order to obtain a general algorithm that can be implemented on different flying microrobots. In the experiments, we employed two flapping-wing flying platforms: the RoboBee, a \num{75}-\si{\milli\gram} two-winged robot, and Bee\textsuperscript{+}, a \num{95}-\si{\milli\gram} four-winged robot. The measured data show that the proposed control method successfully enables both prototypes to hover at desired positions. The data also indicate that to create flying microrobots with the high maneuverability observed in animals such as flies and bees, more advanced designs and fabrication techniques are required to produce better actuation capabilities. Since all the state-of-the-art FWMAVs are underactuated, key elements to be improved are the lift-to-weight ratio and system bandwidth. Another important feature that must be addressed is energy autonomy. All the current functional flying microrobots use piezoelectric actuators and are powered by external energy sources, so are necessarily tethered. A radical new non-electrical actuation method may be required to avoid this problem, as the energy densities in electric batteries are not high enough to generate fully autonomous robots.

\bibliographystyle{IEEEtran}
\bibliography{paper}

\begin{thebibliography}{10}
\providecommand{\url}[1]{#1}
\csname url@samestyle\endcsname
\providecommand{\newblock}{\relax}
\providecommand{\bibinfo}[2]{#2}
\providecommand{\BIBentrySTDinterwordspacing}{\spaceskip=0pt\relax}
\providecommand{\BIBentryALTinterwordstretchfactor}{4}
\providecommand{\BIBentryALTinterwordspacing}{\spaceskip=\fontdimen2\font plus
\BIBentryALTinterwordstretchfactor\fontdimen3\font minus
  \fontdimen4\font\relax}
\providecommand{\BIBforeignlanguage}[2]{{%
\expandafter\ifx\csname l@#1\endcsname\relax
\typeout{** WARNING: IEEEtran.bst: No hyphenation pattern has been}%
\typeout{** loaded for the language `#1'. Using the pattern for}%
\typeout{** the default language instead.}%
\else
\language=\csname l@#1\endcsname
\fi
#2}}
\providecommand{\BIBdecl}{\relax}
\BIBdecl

\bibitem{wood2003microrobotics}
R.~J. Wood, S.~Avadhanula, M.~Menon, and R.~S. Fearing, ``{Microrobotics Using
  Composite Materials: The Micromechanical Flying Insect Thorax},'' in
  \emph{Proc. 2003 IEEE Int. Conf. Robot. Autom.}, {Taipei}, {Taiwan}, Sep.
  2003, pp. 1842--1849.

\bibitem{ma2012design}
K.~Y. Ma, S.~M. Felton, and R.~J. Wood, ``{Design, Fabrication, and Modeling of
  the Split Actuator Microrobotic Bee},'' in \emph{Proc. 2012 IEEE Int. Conf.
  Intell. Robot. Syst.}, {Vilamoura, Algarve}, {Portugal}, Oct. 2012, pp.
  1133--1140.

\bibitem{ma2013controlled}
K.~Y. Ma, P.~Chirarattananon, S.~B. Fuller, and R.~J. Wood, ``{Controlled
  Flight of a Biologically Inspired, Insect-Scale Robot},'' \emph{Science},
  vol. 340, no. 6132, pp. 603--607, May 2013.

\bibitem{yang2019bee}
{X. {Yang} and Y. {Chen} and L. {Chang} and A. A. Calder{\'o}n and N. O.
  {P{\'e}rez-Arancibia}}, ``{Bee\textsuperscript{+}: A 95-mg Four-Winged
  Insect-Scale Flying Robot Driven by Twinned Unimorph Actuators},'' \emph{IEEE
  Robot. Autom. Lett.}, vol.~4, no.~4, pp. 4270--4277, Oct. 2019.

\bibitem{tanaka2005flight}
H.~Tanaka, K.~Hoshino, K.~Matsumoto, and I.~Shimoyama, ``{Flight Dynamics of a
  Butterfly-Type Ornithopter},'' in \emph{Proc. 2005 IEEE/RSJ Int. Intell.
  Robot. Syst.}, {Edmonton, Alta.}, {Canada}, Aug. 2005, pp. 2706--2711.

\bibitem{fearing2000wing}
R.~S. Fearing, K.~H. Chiang, M.~H. Dickinson, D.~Pick, M.~Sitti, and J.~Yan,
  ``{Wing Transmission for a Micromechanical Flying Insect},'' in \emph{Proc.
  2000 IEEE Int. Conf. Robot. Autom.}, {San Francisco, CA}, {USA}, Apr. 2000,
  pp. 1509--1516.

\bibitem{yan2002wing}
J.~Yan, S.~A. Avadhanula, J.~Birch, M.~H. Dickinson, M.~Sitti, T.~Su, and R.~S.
  Fearing, ``{Wing Transmission for a Micromechanical Flying Insect},''
  \emph{J. Micromech.}, vol.~1, no.~3, pp. 221--237, Jun. 2001.

\bibitem{wood2007design}
R.~J. Wood, ``{Design, Fabrication, and Analysis of a 3DOF, 3cm Flapping-Wing
  MAV},'' in \emph{Proc. 2007 IEEE Int. Conf. Intell. Robot. Syst.}, {San
  Diego, CA}, {USA}, Oct. 2007, pp. 1576--1581.

\bibitem{wood2008microrobot}
R.~J. Wood, S.~Avadhanula, R.~Sahai, E.~Steltz, and R.~S. Fearing,
  ``{Microrobot Design Using Fiber Reinforced Composites},'' \emph{J. Mech.
  Des.}, vol. 130, no.~5, p. 052304, May 2008.

\bibitem{perez2011first}
N.~O. P{\'e}rez-Arancibia, K.~Y. Ma, K.~C. Galloway, J.~D. Greenberg, and R.~J.
  Wood, ``{First Controlled Vertical Flight of a Biologically Inspired
  Microrobot},'' \emph{Bioinsp. Biomim.}, vol.~6, no.~3, p. 036009, Aug. 2011.

\bibitem{perez2015model}
N.~O. P{\'e}rez-Arancibia, P.-E.~J. Duhamel, K.~Y. Ma, and R.~J. Wood,
  ``{Model-Free Control of a Hovering Flapping-Wing Microrobot},'' \emph{J.
  Intell. Robot. Syst.}, vol.~77, no.~1, pp. 95--111, Jan. 2015.

\bibitem{chirarattananon2014adaptive}
P.~Chirarattananon, K.~Y. Ma, and R.~J. Wood, ``{Adaptive Control of a
  Millimeter-Scale Flapping-Wing Robot},'' \emph{Bioinsp. Biomim.}, vol.~9,
  no.~2, p. 025004, May 2014.

\bibitem{fuller2019four}
S.~B. Fuller, ``{Four Wings: An Insect-Sized Aerial Robot with Steering Ability
  and Payload Capacity for Autonomy},'' \emph{IEEE Robot. Autom. Lett.},
  vol.~4, no.~2, pp. 570--577, Apr. 2019.

\bibitem{chen2017lyapunov}
Y.~Chen and N.~O. P\'erez-Arancibia, ``{Lyapunov-Based Controller Synthesis and
  Stability Analysis for the Execution of High-Speed Multi-Flip Quadrotor
  Maneuvers},'' in \emph{{Proc. Amer. Control Conf}}, {Seattle, WA}, {USA}, May
  2017, pp. 3599--3606.

\bibitem{murray2017mathematical}
R.~M. Murray, \emph{{A Mathematical Introduction to Robotic
  Manipulation}}.\hskip 1em plus 0.5em minus 0.4em\relax {Boca Raton, FL},
  {USA}: CRC press, 2017.

\bibitem{singer2018clip}
E.~K. Singer, L.~Chang, A.~A. Calder{\'o}n, and N.~O. P{\'e}rez-Arancibia,
  ``{Clip-Brazing for the Design and Fabrication of Micronewton-Resolution
  Millimeter-Scale Force Sensors},'' \emph{Smart Mater. Struct.}, vol.~28,
  no.~5, p. 055028, Apr. 2019.

\bibitem{ma2015mechanical}
K.~Y. Ma, ``{Mechanical Design and Manufacturing of an Insect-Scale
  Flapping-Wing Robot},'' Ph.D. dissertation, Harvard University, 2015.

\bibitem{whitney2010aeromechanics}
J.~P. Whitney and R.~J. Wood, ``{Aeromechanics of Passive Rotation in Flapping
  Flight},'' \emph{J. Fluid Mech.}, vol. 660, pp. 197--220, Oct. 2010.

\bibitem{chang2016dynamics}
L.~Chang and N.~O. P{\'e}rez-Arancibia, ``{The Dynamics of Passive
  Wing-Pitching in Hovering Flight of Flapping Micro Air Vehicles Using
  Three-Dimensional Aerodynamic Simulations},'' in \emph{AIAA Atmosp. Flight
  Mech. Conf.}, {San Diego, CA}, {USA}, Jan. 2016, p. 0013.

\bibitem{chang2018time}
------, ``{Time-Averaged Dynamic Modeling of a Flapping-Wing Micro Air Vehicle
  with Passive Rotation Mechanisms},'' in \emph{Atmosp. Flight Mech. Conf.},
  {Atlanta, GA}, {USA}, Jun. 2018, p. 2830.

\bibitem{wood2008first}
R.~J. Wood, ``{The First Takeoff of a Biologically Inspired At-Scale Robotic
  Insect},'' \emph{IEEE Trans. Robot.}, vol.~24, no.~2, pp. 341--347, 2008.

\bibitem{finio2012open}
B.~M. Finio and R.~J. Wood, ``{Open-Loop Roll, Pitch and Yaw Torques for a
  Robotic Bee},'' in \emph{Proc.2012 IEEE Int. Conf. Intell. Robot. Syst.},
  {Vilamoura, Algarve}, {Portugal}, Oct. 2012, pp. 113--119.

\bibitem{gravish2016anomalous}
N.~Gravish and R.~J. Wood, ``{Anomalous Yaw Torque Generation from Passively
  Pitching Wings},'' in \emph{Proc. 2016 IEEE Int. Conf. Robot. Autom.},
  {Stockholm}, {Sweden}, May 2016, pp. 3282--3287.

\bibitem{Ying2016ICRA}
Y.~Chen and N.~O. P\'erez-Arancibia, ``{Generation and Real-Time Implementation
  of High-Speed Controlled Maneuvers Using an Autonomous 19-Gram Quadrotor},''
  in \emph{{Proc. 2016 IEEE Int. Conf. Robot. Autom.}}, {Stockholm}, {Sweden},
  May 2016, pp. 3204--3211.

\bibitem{roberts2011adaptive}
A.~Roberts and A.~Tayebi, ``{Adaptive Position Tracking of VTOL UAVs},''
  \emph{IEEE Trans. Robot.}, vol.~27, no.~1, pp. 129--142, Feb. 2011.

\bibitem{Ying2018IROS}
Y.~Chen and N.~O. P\'erez-Arancibia, ``{Nonlinear Adaptive Control of Quadrotor
  Multi-Flipping Maneuvers in the Presence of Time-Varying Torque Latency},''
  in \emph{{Proc. 2018 IEEE Int. Conf. Intell. Robot. Syst.}}, {Madrid},
  {Spain}, Oct. 2018, pp. 1--9.

\end{thebibliography}
\balance

\end{document}